\documentclass[letterpaper, 10 pt, journal, twoside]{IEEEtran} 
\usepackage{caption}
\usepackage{algorithm}
\usepackage{algpseudocode}

\usepackage{booktabs} 
\usepackage{verbatim}
\usepackage{graphicx} 
\usepackage{makecell} 
\usepackage{amsmath} 
\usepackage{amssymb} 
\usepackage{xcolor,blindtext} 
\usepackage{subfig}
\usepackage{color} 
\usepackage{balance} 
\usepackage[sort&compress,square,comma,numbers]{natbib} 

\DeclareMathOperator*{\argmin}{arg\,min}

\begin{document}

\title{Optical flow-based vascular respiratory motion compensation}

\author{Keke Yang, Zheng Zhang, Meng Li, Tuoyu Cao, Maani Ghaffari, and Jingwei Song*%
	\thanks{K. Yang, Z. Zhang, M. Li, and T. Cao  are with United Imaging, Shanghai, China. \texttt{\{keke.yang, zheng.zhang, meng.li02, tuoyu.cao\}@united-imaging.com}. }
   \thanks{M. Ghaffari is with the University of Michigan, Ann Arbor, MI 48109, USA. \texttt{maanigj@umich.edu}.}%

   \thanks{J. Song* (Corresponding author) is with United Imaging Research Institute of Intelligent Imaging, Beijing 100144, China
 \texttt{jingweisong.eng@outlook.com}}%
}
\maketitle

\begin{abstract}

This paper develops a new vascular respiratory motion compensation algorithm, Motion-Related Compensation (MRC), to conduct vascular respiratory motion compensation by extrapolating the correlation between invisible vascular and visible non-vascular. Robot-assisted vascular intervention can significantly reduce the radiation exposure of surgeons. In robot-assisted image-guided intervention, blood vessels are constantly moving/deforming due to respiration, and they are invisible in the X-ray images unless contrast agents are injected. The vascular respiratory motion compensation technique predicts 2D vascular roadmaps in live X-ray images. When blood vessels are visible after contrast agents injection, vascular respiratory motion compensation is conducted based on the sparse Lucas-Kanade feature tracker. An MRC model is trained to learn the correlation between vascular and non-vascular motions. During the intervention, the invisible blood vessels are predicted with visible tissues and the trained MRC model. Moreover, a Gaussian-based outlier filter is adopted for refinement. Experiments on in-vivo data sets show that the proposed method can yield vascular respiratory motion compensation in $0.032 \sec$, with an average error $1.086~\mathrm{mm}$. Our real-time and accurate vascular respiratory motion compensation approach contributes to modern vascular intervention and surgical robots.
\end{abstract}


\IEEEpeerreviewmaketitle

\section{Introduction}

Robot-assisted vascular interventional therapy is a rapidly developing technology in the field of cardiovascular disease treatment~\cite{rossle2013tips,tsurusaki2015surgical,ambrosini2015continuous}. Its value in peripheral blood vessels, particularly in tumor embolization therapy, is gaining increasing attention. Among them, robot-assisted vascular intervention reduces radiation and has drawn more attention recently~\cite{duan2023technical}. In the vascular intervention procedure, roadmapping is the process of superimposing 2D vascular on live fluoroscopic images and plays a key role in surgeries~\cite{riviere2006robotic}.\par

\begin{figure}[t]
    \centering
    \includegraphics[width=0.99\columnwidth]{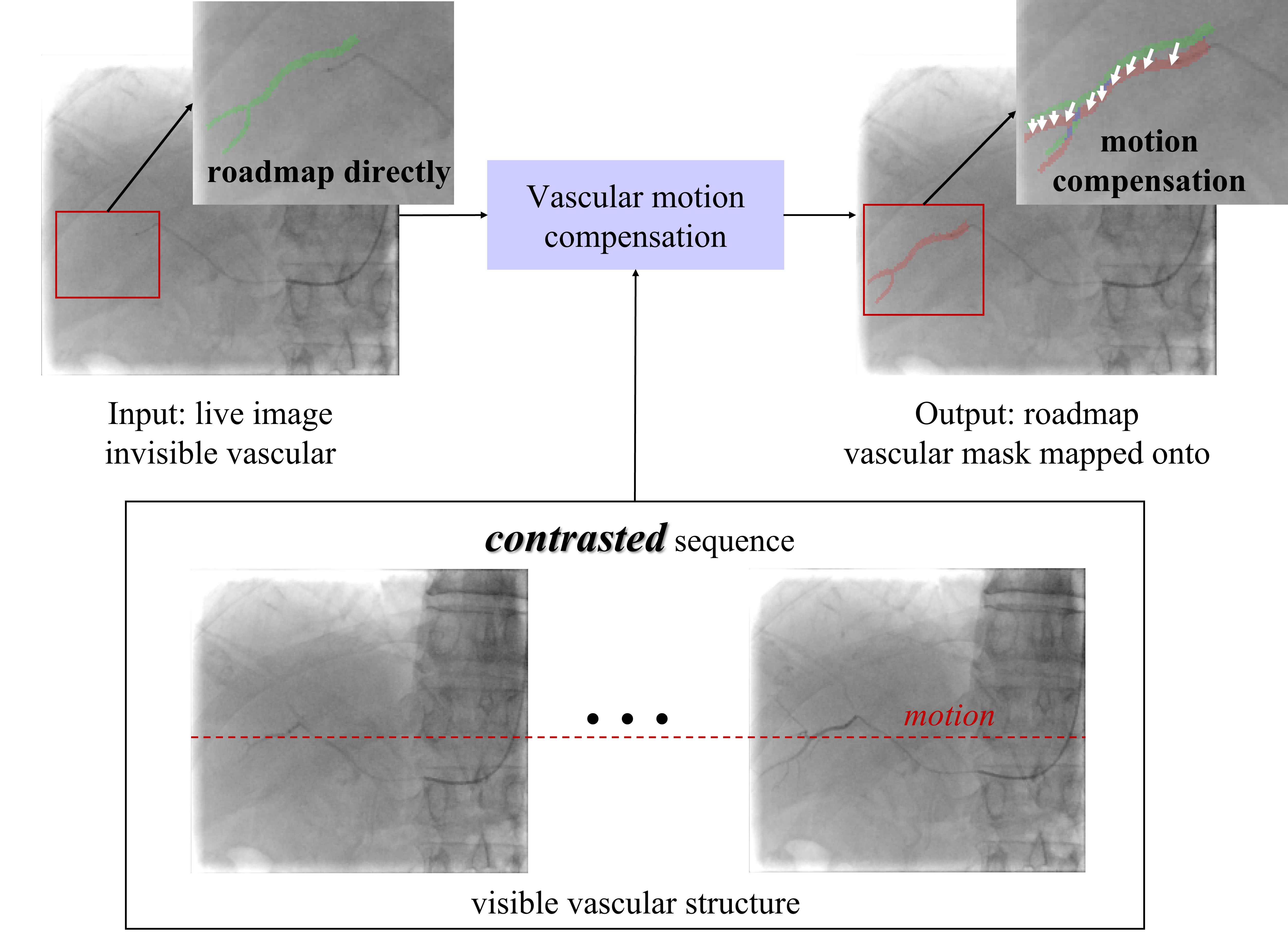}
    \caption{The scenario of vascular respiratory motion compensation: input live image with invisible vascular, output roadmap of the live image with mapped red vascular mask. The roadmap can be obtained by contrasted sequences and mapped onto the live image directly, as indicated by the green mask. Vascular respiratory motion compensation utilizes contrasted sequences to predict motion between the live image and roadmap, as white arrows indicate. The red dotted line is for easy visualization of motion.}
    \label{fig:scene}
\end{figure}

Vascular respiratory motion compensation (or dynamic roadmapping) technique pushes conventional \textbf{static} roadmaps to \textbf{dynamic} roadmaps and helps the interventionists/robots manipulate catheters and guidewires or place stents by visualization of the map and devices on one screen~\cite{riviere2006robotic}. In typical interventions, catheters and guidewires are guided under live fluoroscopic images, which contain 2D device information~\cite{unger2008image}. During the process, contrast agents are injected to provide clear vascular structures~\cite{wagner2021real}. However, contrast agents flow quickly, and the vascular no longer develops after the contrast agents disappear~\cite{ambrosini2015continuous}. Moreover, deforming/moving roadmapping brings large errors in organs like the liver due to respiration motion and requires additional compensation. Fig.~\ref{fig:scene} shows a typical vascular  respiratory motion compensation for handling the two issues in conventional static roadmapping. The green mask obtained from the contrasted image (the fluoroscopic image with contrast agent injection as shown in Fig.~\ref{fig:scene} contrasted sequences) is mapped onto the live image directly. The error is evident from the guidewire due to breathing motion. Vascular respiratory motion compensation can dynamically move 2D roadmaps to properly match the live fluoroscopic images, especially after the contrast agent disappears and vascular structures are not visible from fluoroscopic images. The red mask in Fig.~\ref{fig:scene} is the prediction of vascular respiratory motion compensation, which can provide immediate feedback to robots/physicians during surgeries.

Existing vascular respiratory motion compensation methods can be categorized as respiratory state-driven and motion modeling~\cite{wagner2021real}. Compensation based on respiratory state extraction connects vascular motion with the respiratory state, which can be extracted through external devices or images. External devices include respiratory belts~\cite{giraud2013respiratory}, surface markers~\cite{spinczyk2014methods, sayeh2007respiratory}, and electromagnetic sensors~\cite{loschak2020automatically}. Although estimating the respiratory state with external devices-based is straightforward and applicable, it disrupts the clinical workflow and is not robust due to unfixed respiratory rate and amplitude~\cite{wagner2021real}. Unlike using additional devices, image-based approaches estimate respiratory state with anatomical landmarks such as the diaphragm to obtain respiratory state~\cite{wagner2021real, wagner2017feature, king2009subject, fischer2016unsupervised}. \cite{wagner2021real} estimated respiratory state based on diaphragm location in the non-contrasted images and fitted  between vascular affine transformation and respiratory state with a linear function. For non-contrasted live images, an affine transformation of vascular was estimated by respiratory state, which is computationally efficient. \cite{wagner2021real} reported an average one-frame processing time of $17~ms$, which is the fastest. Image-based respiratory state extraction requires no additional clinical workflow but is limited by the Field-of-View (FoV). As~\cite{henningsson2013advanced} pointed out, although respiratory state-based methods are time-efficient, their accuracy and robustness are limited because motions over the respiratory cycle are generally less reproducible.

Different from respiratory state-driven approaches, model-based approaches build models to learn and predict the motion in fluoroscopic frames~\cite{vernikouskaya2022deep, ma2020dynamic, ambrosini2015continuous,
atasoy2008real, orozco2008respiratory}.  Model-based methods can be categorized as \textbf{catheter-based} and \textbf{catheter-free}. \textbf{Catheter-based} methods track the catheter tip to conduct vascular respiratory motion compensation.
\cite{vernikouskaya2022deep} learned displacements of the catheter tip by a Convolutional Neural Network (CNN). In their works, vascular motion was factorized into respiratory and heart motion. Heart motion compensation was done by ECG, and respiratory motion was predicted by CNN. \cite{ma2020dynamic} conducted cardiac and respiratory-related vascular motion compensation by ECG alignments and catheter tip tracking in X-ray fluoroscopy, respectively. In particular, to realize accurate and robust tracking of the catheter tip, they proposed a new deep Bayesian filtering method that integrated the detection outcome of a CNN and the sequential motion estimation using a particle filtering framework.
\textbf{Catheter-free} methods conduct vascular respiratory motion compensation by soft tissue motion. \cite{zhu2010image, manhart2011self} utilized soft tissue around the heart to model vascular motion. These works are based on the assumption that vascular motion and soft tissue motion followed the same affine transformation, and Lucas-Kanade (LK) tracker was improved to estimate soft tissue affine transformation to conduct vascular respiratory motion compensation. ~\cite{zhu2010image} proposed special handling of static structures to recover soft tissue motion. ~\cite{zhu2010image} also applied the LK tracker on multiple observed fluoroscopic images to gain robustness. However, the LK tracker on the entire X-ray image needs heavy computation. Meanwhile, vascular motion and soft tissue motion are not always consensus. To sum up, methods based on motion modeling can yield high accuracy and robust compensations without additional input, but their computations are large. To our knowledge, no study can achieve both real-time implementation and high accuracy in hepatic vascular respiratory motion compensation.

In this paper, we propose Motion-Related Compensation (MRC) algorithm to achieve real-time and accurate vascular respiratory motion compensation. To enable fast vascular respiratory motion compensation, feature points are extracted from the observed fluoroscopic image; deforming vascular motions are estimated based on the tracked feature points. Furthermore, a correlation model is built between vascular points motion and non-vascular points motion when the contrast agent develops in X-ray images. After the disappearance of the contrast agent, our trained model can predict the invisible motion of the vascular points based on the motion of the non-vascular points. Moreover, a Gaussian-based Outlier Filtering (GOF) technique is adopted to refine the correlation model's prediction. In summary, our main contributions are:
\begin{itemize}
    \item To our knowledge, the proposed MRC is the first applicable method that is both real-time and accurate for respiratory motion compensation. It  achieves $31 Hz$ and $1.086 mm$ for the typical fluoroscopic image size $512 \times 512$ on a modern desktop with Intel Core i5-10500 CPU at 3.10GHz.
    \item We propose a novel vascular MRC method without assuming that vascular and non-vascular motion are identical, which uses multi-frame contrasted images to learn the model of vascular-nonvascular motion correlation.
    \item GOF is adopted to improve the accuracy of our MRC predictions.
\end{itemize}

\section{Preliminaries}
\label{sec_preliminary}

Fig.~\ref{fig:scene} describes the vascular respiratory motion compensation process. Blood vessels are visible from a small sequence with the contrast agent. After the contrast agent flows, the vascular is not visible in the live X-ray images. Vascular respiratory motion compensation on live images significantly benefits surgeons and surgical robots. Thus, the purpose of this research is to implement vascular motion compensation on live X-ray images based on the limited images with visible vascular structures.

Denote the sequence with and without the contrast agent as 2D images $\mathbf{I} = \left\{\mathbf{I}_{1}, \mathbf{I}_{2},..., \mathbf{I}_{k}\right\}$ and $\mathbf{R}= \left\{\mathbf{R}_{1}, \mathbf{R}_{2},..., \mathbf{R}_{q}, ...\right\}$. Reference frame taken from contrasted images is specially defined as $\mathbf{I}_{r}$. The binary vascular mask of reference frame $\mathbf{I}_{r}$ is denoted as $\mathbf{M}_{r}$. The vascular corners and non-vascular corners extracted from reference frame $\mathbf{I}_{r}$ by Shi-Tomasi approach~\cite{shi1994good}  are denoted as $\mathbf{C}_{r}^{v} = [\mathbf{c}_{r}^{v(1)},...,\mathbf{c}_{r}^{v(\mathrm{N}_{v})}]^\top$ and $\mathbf{C}_{r}^{n} = [\mathbf{c}_{r}^{n(1)},...,\mathbf{c}_{r}^{n(\mathrm{N}_{n})}]^\top$, where $\mathbf{c}_{r}^{v(i)},\mathbf{c}_{r}^{n(i)} \in \mathbb{R}^{2 \times 1} $ and $\mathrm{N}_{v}$, $\mathrm{N}_{n}$ are the number of vascular corners and non-vascular corners. The vascular corners motion flow between $\mathbf{I}_{i}$ and $\mathbf{I}_{r}$ is denoted as $\mathbf{D}_{i}^{v}=[\mathbf{d}_{i}^{v(1)}, ..., \mathbf{d}_{i}^{v(\mathrm{N}_{v})}]^\top \in \mathbb{R}^{\mathrm{N}_{v} \times 2}$, where $\mathbf{d}_{i}^{v(j)} \in \mathbb{R}^{2 \times 1}$. The non-vascular motion flow is denoted as $\mathbf{D}_{i}^{n} = [\mathbf{d}_{i}^{n(1)}, ..., \mathbf{d}_{i}^{n(\mathrm{N}_{n})}]^\top \in \mathbb{R}^{\mathrm{N}_{n} \times 2}$, where $\mathbf{d}_{i}^{n(j)} \in \mathbb{R}^{2 \times 1}$. The vascular corners motion flow between $\mathbf{R}_{q}$ and $\mathbf{I}_{r}$ is denoted as $\mathbf{F}_{q}^{v}=[\mathbf{f}_{q}^{v(1)}, ..., \mathbf{f}_{q}^{v(\mathrm{N}_{v})}]^\top \in \mathbb{R}^{\mathrm{N}_{v} \times 2}$, where $\mathbf{f}_{q}^{v(j)} \in \mathbb{R}^{2 \times 1}$. The non-vascular motion flow is denoted as $\mathbf{F}_{q}^{n} = [\mathbf{f}_{q}^{n(1)}, ..., \mathbf{f}_{q}^{n(\mathrm{N}_{n})}]^\top \in \mathbb{R}^{\mathrm{N}_{n} \times 2}$, where $\mathbf{f}_{q}^{n(j)} \in \mathbb{R}^{2 \times 1}$.

\section{Method}

\subsection{System Overview}

\begin{figure*}[!ht]
    \centering
    \includegraphics[width=13cm]{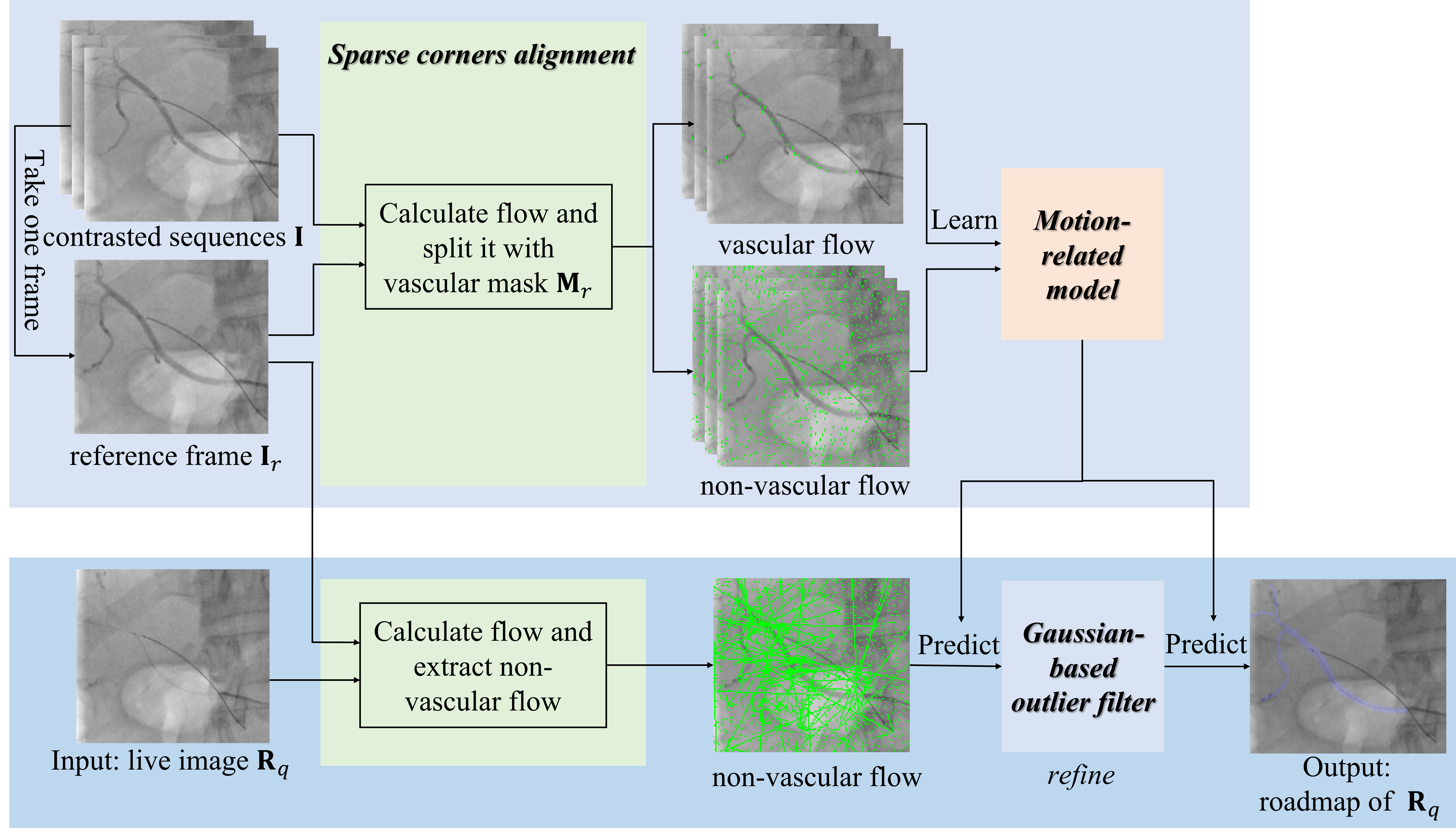}
    \caption{Overview of the proposed MRC algorithm.}
    \label{fig:algorithm}
\end{figure*}

Fig.~\ref{fig:algorithm} shows the pipeline of our proposed MRC algorithm consisting of three modules: \textbf{sparse corners alignment}, \textbf{motion-related model}, and \textbf{GOF}. \textbf{Sparse corners alignment} module calculates motion flow between reference frame $\mathbf{I}_{r}$ and live moving frames and splits motion flow into vascular motion flow and non-vascular motion flow with vascular mask $\mathbf{M}_{r}$ extracted from the reference frame. \textbf{Motion-related model} builds the correlation between vascular motion flow and non-vascular motion flow. \textbf{GOF} module filters outliers with the obtained non-vascular motion flow to refine prediction.

\subsection{Sparse Corners Alignment}

Affine motion parameterization is used in~\cite{zhu2010image, manhart2011self} to model non-vascular motion incurred by respiration, which is regarded as vascular motion directly. Affine motion parameterization is also used in ~\cite{wagner2021real} to model vascular motion. However, the affine motion model may not simulate real vascular motion due to a low degree of freedom. Moreover, it is not accurate to treat vascular motion and non-vascular motion equally; for example, the amplitude of respiratory motion at the top of the liver is larger than at the bottom of the liver. Therefore, we adopt sparse optical flow as a tracker to obtain motion at the pixel level, which does not limit the freedom of motion nor assume the consistency of vascular motion and non-vascular motion. Sparse flow selects a sparse feature set of pixels (e.g., interesting features such as edges and corners) to track its motion vector, which only tracks much fewer points but achieves fast speed and high accuracy. Inspired by ~\cite{liu2022rgb}, Shi-Tomasi corner detection method~\cite{shi1994good} is used to extract sparse features, and LK~\cite{bouguet2001pyramidal} is used to track corners' motion sequence in this paper. And experiments in Section IV validate their efficiencies.

Sparse corner alignment calculates vascular and non-vascular motion flow separately. Firstly, the vascular corners $\mathbf{C}_{r}^{v} \in \mathbb{R}^{\mathrm{N}_{v} \times 2}$ and the non-vascular corners $\mathbf{C}_{r}^{n} \in \mathbb{R}^{\mathrm{N}_{n} \times 2}$ in reference frame $\mathbf{I}_{r}$ are extracted using Shi-Tomasi~\cite{shi1994good}. Then, sparse vascular motion flow $\mathbf{D}_{i}^{v} \in \mathbb{R}^{\mathrm{N}_{v} \times 2}$ and non-vascular motion flow $\mathbf{D}_{i}^{n} \in \mathbb{R}^{\mathrm{N}_{n} \times 2}$ between other frame $\mathbf{I}_{i}$ and $\mathbf{I}_{r}$ can be estimated by  LK~\cite{bouguet2001pyramidal}. Similarly, for any live image $\mathbf{R}_{q}$, its non-vascular motion flow $\mathbf{F}_{q}^{n}$ can be obtained by LK~\cite{bouguet2001pyramidal}.

Although dense flow can also obtain motion at the pixel level, it computes the optical flow sequence for every pixel, which produces accurate results but runs much slower. In addition, ORB and SIFT are more efficient in aligning sparse features with scale and orientation differences. Experiments show that ORB and SIFT are less robust in vascular respiratory motion compensation.

\subsection{Motion-Related Model}
\label{model}

The motion-related model first formulates a correlation between vascular and non-vascular motion and then predicts vascular motion after contrast agents disappear. We assume the motion of soft tissues caused by breathing is smooth, for example, the hepatic artery follows the diaphragm to make the correlated motion. Inspired by this, we assume that vascular motion is caused by the motion of the surrounding non-vascular tissues. Enforcing the smoothness presumption, our motion-related model retrieves vascular motion and non-vascular motion based on the sequence with visible vascular structures. It should be noted that the model parameters need to be updated for each injection of the contrast agent for each patient. Our model can only carry out vascular respiratory motion compensation for the blood vessels that has been observed. It is reasonable to do vascular respiratory motion compensation by the peripheral non-vascular motion that is elastic and connected to blood vessels~\cite{fujimoto2021estimation}. It should be noted that contrast agents flow quickly, and our algorithm can reduce the use of contrast agents.

\cite{zhu2010image} retrieves  vascular respiratory motion considering vascular and non-vascular motion as the same. It is not accurate in practice due to different motion magnitudes incurred by respiration. To establish a relation between vascular and non-vascular motion, a linear and a non-linear regression can be used to fit the relation. Its slow speed~\cite{kabzan2019learning}, hyperparameters sensitivity, and similar accuracy to linear regression drive us to select a linear model. In addition, ~\cite{fujimoto2021estimation} used the linear elastic model, which indicates a linear regression. And experiments also verify the effectiveness of the linear regression model.

Specifically, the Pearson coefficient is adopted to quantify the correlation~\cite{Statistics}. The $i$th vascular corner motion flows on $\mathrm{k}$ frames are denoted as $\mathbf{Y}_{i} = [\mathbf{d}_{1}^{v(i)}, ..., \mathbf{d}_{k}^{v(i)}]\in \mathbb{R}^{2 \times k}$. The $j$th non-vascular motion flows on $\mathrm{k}$ frames are denoted as $\mathbf{X}_{j} = [\mathbf{d}^{n(j)}_{1}, ..., \mathbf{d}^{n(j)}_{k}]\in \mathbb{R}^{2 \times k}$. Pearson coefficient between the ${i}$th vascular corner motion flows, and the ${j}$th non-vascular motion flows can be calculated by

\begin{equation}
    \rho_{i,j} = \frac{\operatorname{cov}(\mathbf{X}_{j}|_{x}, \mathbf{Y}_{i}|_{x})}{\sigma_{\mathbf{X}_{j}|_{x}} \sigma_{\mathbf{Y}_{i}|_{x}}} \cdot \frac{\operatorname{cov}(\mathbf{X}_{j}|_{y}, \mathbf{Y}_{i}|_{y})}{\sigma_{\mathbf{X}_{j}|_{y}} \sigma_{\mathbf{Y}_{i}|_{y}}},
    \label{Pearson correlation}
\end{equation}

\noindent where $|_{x}$ represents its first component and $|_{y}$ represents its second component, $\operatorname{cov}(\cdot,\cdot)$ calculates the covariance between two vectors. The $j$th non-vascular corner is used to predict the $i$th vascular corner motion flow if $\rho_{i,j} > \rho_{th}$ where $\rho_{th}$ is a predefined threshold. Then, least square serves as linear regressor between $\mathbf{X}_{j}$ and $\mathbf{Y}_{i}$ as \eqref{fitting} shows. And $\rho$ and $\hat{\mathbf{Q}}$ are correlation model parameters, which can be used to predict vascular corner motion flow by non-vascular motion flow. 

\begin{equation}
\begin{aligned}
    \mathbf{\hat{Q}}_{i,j} &= 
     \begin{bmatrix}
    \hat{\mathrm{a}}_{x} & \hat{\mathrm{a}}_{y} \\
    \hat{\mathrm{b}}_{x} &\hat{\mathrm{b}}_{y} 
    \end{bmatrix}\\
    \hat{\mathrm{a}}_{x}, \hat{\mathrm{b}}_{x} &= \argmin \limits_{\mathrm{a}, \mathrm{b}} \sum_{m=1}^{\mathrm{k}}\Vert\mathrm{a} \cdot \mathbf{d}_{m}^{n(j)}|_{x} + \mathrm{b} - \mathbf{d}_{m}^{v(i)}|_{x}\Vert^2  \\
    \hat{\mathrm{a}}_{y}, \hat{\mathrm{b}}_{y} &= \argmin \limits_{\mathrm{a}, \mathrm{b}} \sum_{m=1}^{\mathrm{k}}\Vert\mathrm{a} \cdot \mathbf{d}_{m}^{n(j)}|_{y} + \mathrm{b} - \mathbf{d}_{m}^{v(i)}|_{y}\Vert^2 
\end{aligned},
    \label{fitting}
\end{equation}

In summary, for each vascular corner, $\mathbf{c}_{r}^{v(i)}$ in $\mathbf{I}_{r}$, non-vascular points with high correlation are selected based on pre-defined threshold $\rho_{th}$. Then, fitting parameters between this vascular corner and each selected non-vascular corner point are calculated by \eqref{fitting}. The procedure of establishing the motion-related model is shown in Algorithm \ref{sparse training}. 

\begin{algorithm}
    \small 
    \caption{Motion-related model estimation process}
    \label{sparse training}
    \begin{algorithmic}[1]
    \Require  reference frame $\mathbf{I}_{r}$, contrasted sequences $\mathbf{I}$

    \Ensure weight matrix $\mathbf{W}\in \mathbb{R}^{\mathrm{N}_{v} \times \mathrm{N}_{n}}$ and fitting parameters matrix $\mathbf{L}^{A},\mathbf{L}^{B}\in \mathbb{R}^{\mathrm{N}_{v} \times \mathrm{N}_{n} \times 2}$
    
    \State Initialize $\mathbf{W}$ and $\mathbf{L}^{A},\mathbf{L}^{B}$ with zeros matrix
   
    \State Calculate training vascular motion flows $ \{ \mathbf{D}_{1}^{v}, \mathbf{D}_{2}^{v}, ..., \mathbf{D}_{k}^{v} \}$ and training non-vascular motion flows $ \{ \mathbf{D}_{1}^{n}, \mathbf{D}_{2}^{n}, ..., \mathbf{D}_{k}^{n} \}$
    
    \For{each vascular corner $i \in [1, \mathrm{N}_{v}]$}
    \For{each non-vascular corner $j \in [1, \mathrm{N}_{n}]$}
    \State Calculate Pearson coefficient $\rho_{i,j}$ by \eqref{Pearson correlation}
    \If{$\rho_{i,j} > \rho_{th}$}
    \State $\mathbf{W}_{i,j} = \rho_{i,j}$
    \State Calculate fitting parameters $\hat{\mathbf{Q}}_{i,j}$ by \eqref{fitting} 
    \State $\mathbf{L}^{A}_{i,j,.}=[\hat{\mathrm{a}}_{x}, \hat{\mathrm{a}}_{y}]$, $\mathbf{L}^{B}_{i,j,.}=[\hat{\mathrm{b}}_{x}, \hat{\mathrm{b}}_{y}]$

    \EndIf

    \EndFor

    \State Normalize weight vector $\mathbf{W}_{i,.} = \frac{\mathbf{W}_{i,.}}{\sum_{j=1}^{\mathrm{N}_{n}}\mathbf{W}_{i,j}}$ 
    \EndFor
        
    \end{algorithmic}
    Notion: the for loop is implemented in vectorized form for efficiency.
\end{algorithm}

The $i$th vascular corner motion flow is predicted by the weighted average of all selected non-vascular motion flows according to
\begin{equation}
    \hat{\mathbf{f}}_{q}^{v(i)} = (\mathbf{W}_{i,.} \cdot (\mathbf{L}^{A}_{i,.,.} \odot \mathbf{F}^{n}_{q} + \mathbf{L}^{B}_{i,.,.}))^\top,
    \label{weighting}
\end{equation}

\noindent where $\odot$ denotes Hadamard product (element-wise multiplication), $\hat{\mathbf{f}}_{q}^{v(i)}\in \mathbb{R}^{2 \times 1}$ is the predicted $i$th vascular corner motion flow between $\mathbf{R}_{q}$ and $\mathbf{I}_{r}$. However, sparse motion flows of some non-vascular corner points between $\mathbf{R}_{q}$ and $\mathbf{I}_{r}$ may have large errors, which makes the predicted vascular motion flow $\hat{\mathbf{f}}_{q}^{v(i)}$ sometimes inaccurate. To refine vascular motion flow prediction, we propose GOF-based flow motion prediction in the following subsection.

\subsection{GOF-based Motion Flow Predicting}

GOF-based process refines vascular motion prediction $\hat{\mathbf{f}}_{q}^{v(i)}$ from the motion-related model and deletes outliers based on Gaussian distribution. In order to reduce the influence of large errors between $\mathbf{R}_{q}$ and $\mathbf{I}_{r}$, we assume the predictions are i.i.d and follow Gaussian distribution. 

For the $i$th vascular corner, with fitting coefficients $\mathbf{L}^{A}$ and $\mathbf{L}^{B}$, its non-vascular prediction $\hat{\mathbf{P}}_{i} = [\hat{\mathbf{p}}^{(1)}_{i}, \hat{\mathbf{p}}^{(2)}_{i}, ..., \hat{\mathbf{p}}^{(\mathrm{N}_{n})}_{i}]^\top \in \mathbb{R}^{\mathrm{N}_{n} \times 2}$ is inferred by

\begin{equation}
    \hat{\mathbf{P}}_{i} = \mathbf{L}^{A}_{i,.,.} \odot \mathbf{F}^{n}_{q} + \mathbf{L}^{B}_{i,.,.},
    \label{predict}
\end{equation}

\noindent where $\hat{\mathbf{p}}^{(j)}_{i} \in \mathbb{R}^{2 \times 1}$ represents the $j$th non-vascular prediction for the $i$th vascular corner. Ideally, for any $j \in [1, \mathrm{N}_{n}]$, the value of $\hat{\mathbf{p}}^{(j)}_{i}$ should be equivalent since they belong to the same vascular point. Therefore, each element in $\mathbf{P}_{i}$ should obey Gaussian distribution in both directions. That is variable $\mathbf{P}_{i}|_{x} \sim \mathcal{N}(\mu_{i}|_{x}, \sigma_{i}^{2}|_{x})$, $\mathbf{P}_{i}|_{y} \sim \mathcal{N}(\mu_{i}|_{y}, \sigma_{i}^{2}|_{y})$ where $\mu_{i},\sigma_{i} \in \mathbb{R}^{2}$ are mean and standard deviation of the $i$th vascular corner predicting $ \hat{\mathbf{P}}_{i}$. $\mu_{i}$ and $\sigma_{i}$ can be statistically calculated by \eqref{gauss}. For a random variable $\mathrm{Z} \sim \mathcal{N}(\mu, \sigma^{2})$, it has a probability 0.9974 within the range of $(\mu-3\sigma, \mu+3\sigma)$. Therefore, the outlier in $\hat{\mathbf{P}}_{i}$ caused by non-vascular motion flow estimating error can be deleted based on the $3\sigma$ bound, which updates motion-related model $\mathbf{W}$ as shown in Algorithm~\ref{test}. Then each vascular corner motion flow prediction can be refined by \eqref{refine weighting} utilizing updated weight $\Tilde{\mathbf{W}}$. Algorithm \ref{test} describes how to predict vascular motion flows based on GOF, which outputs vascular motion flows $\Tilde{\mathbf{F}}_{q}^{v} \in \mathbb{R}^{\mathrm{N}_{v} \times 2}$. Finally, the reference frame vascular mask $\mathbf{M}_{r}$ is mapped onto the live image $\mathbf{R}_{q}$ according to the predicted vascular motion flows $\Tilde{\mathbf{F}}_{q}^{v}$.

\begin{equation}
    \begin{aligned}
        \mu_{i}|_{x} &= \frac{\sum_{j=1}^{\mathrm{N}_{n}}\hat{\mathbf{p}}_{i}^{(j)}|_{x}}
        {\mathrm{N}_{n}}, \sigma_{i}|_{x} = \sqrt{\frac{\sum_{j=1}^{\mathrm{N}_{n}}(\hat{\mathbf{p}}_{i}^{(j)}|_{x}-\mu_{i}|_{x})^2}{\mathrm{N}_{n}}},  \\
        \mu_{i}|_{y} &= \frac{\sum_{j=1}^{\mathrm{N}_{n}}\hat{\mathbf{p}}_{i}^{(j)}|_{y}}
        {\mathrm{N}_{n}}, \sigma_{i}|_{y} = \sqrt{\frac{\sum_{j=1}^{\mathrm{N}_{n}}(\hat{\mathbf{p}}_{i}^{(j)}|_{y}-\mu_{i}|_{y})^2}{\mathrm{N}_{n}}},
    \end{aligned}
    \label{gauss}
\end{equation}

\begin{equation}
    \Tilde{\mathbf{f}}_{q}^{v(i)} = (\Tilde{\mathbf{W}}_{i,.} \cdot \hat{\mathbf{P}}_{i})^\top.
    \label{refine weighting}
\end{equation}

\begin{algorithm}
    \small
    \caption{Predicting vascular motion flows based on GOF.}
    \label{test}
    \begin{algorithmic}[1]
        \Require
        weight matrix $\mathbf{W}$, linear fitting coefficient $\mathbf{L}^{A},\mathbf{L}^{B}$, live image $\mathbf{R}_{q}$, reference frame $\mathbf{I}_{r}$
        
        \Ensure
        predicted vascular motion flows $\Tilde{\mathbf{F}}_{q}^{v} \in \mathbb{R}^{\mathrm{N}_{v} \times 2}$ between $\mathbf{R}_{q}$ and $\mathbf{I}_{r}$

        \State Calculate non-vascular motion flows $\mathbf{F}_{q}^{n} \in \mathbb{R}^{\mathrm{N}_{n} \times 2}$ between $\mathbf{R}_{q}$ and $\mathbf{I}_{r}$

        \For{each vascular corner $i \in [1, \mathrm{N}_{v}]$}
        \State Calculate non-vascular predicting $\hat{\mathbf{P}}_{i}$ by \eqref{predict} 
        \State /*        Delete outlier      */ 
        \State Calculate $\mu_{i}, \sigma_{i}$ by \eqref{gauss}
        \For{each non-vascular predicting $j \in [1, \mathrm{N}_{n}]$}
        \If{$\hat{\mathbf{p}}_{i}^{(j)}|_{x} \notin (\mu_{i}|_{x}-3\sigma_{i}|_{x}, \mu_{i}|_{x}+3\sigma_{i}|_{x})$ or $\hat{\mathbf{p}}_{i}^{(j)}|_{y} \notin (\mu_{i}|_{y}-3\sigma_{i}|_{y}, \mu_{i}|_{y}+3\sigma_{i}|_{y})$}
        \State $\mathbf{W}_{i,j} = 0$
        \EndIf
        \EndFor
        \State Re-normalize weight $\Tilde{\mathbf{W}}_{i,.} = \frac{\mathbf{W}_{i,.}}{\sum_{j=1}^{\mathrm{N}_{n}}\mathbf{W}_{i,j}}$ 
        \State Predict vascular motion flow $\Tilde{\mathbf{f}}_{q}^{v(i)}$ by \eqref{refine weighting}
        \EndFor
  
    \end{algorithmic}
    
\end{algorithm}

\section{Experiment Results}

\subsection{Experimental Setup}

To validate our proposed MRC algorithm, X-ray sequences generated during hepatic vascular surgeries TACE and TIPS from Zhongshan Hospital were collected, and we also obtained X-ray image sequences generated from a male porcine.
13 image sequences are screened with contrast agents.
These image sequences include frames with and without contrast agents. During the process, the patient breathed freely. The images are with the size either in $512\times512$ or in $1024\times1024$ and the pixel resolution either $0.746mm$, $0.308mm$, or $0.390mm$. The Region of Interest of the image ranges from $216 \times 231$ to $562 \times 726$. For the images without the contrast agent, physicians \footnote{The second and third authors obtained M.D. degrees and did the score.} manually labeled the vascular centerlines for each sequence, which was used as \textit{the reference} for validation. We performed the proposed MRC on a total of $520$ frames and quantitative evaluation on $222$ labeled frames. Table \ref{tab:data} lists detailed information on 13 sequences used for MRC. We also collected X-ray image sequences generated during coronary artery surgeries, but we did not count them due to poor image quality and large deformation.

\begin{table}[]
    \renewcommand{\arraystretch}{1.} 
    \setlength{\tabcolsep}{0.1pt} 
    \caption{Number of images used to compensate vascular motion. Seq.A-Seq.M represents different sequences.}
    \label{tab:data}
    \centering

    \scriptsize
    \begin{tabular}{l c c c c c c c c c c c c c}
    \toprule
   
     & Seq.A & Seq.B & Seq.C & Seq.D & Seq.E & Seq.F & Seq.G & Seq.H & Seq.I & Seq.J & Seq.K & Seq.L & Seq.M\\

    \midrule
    \makecell[l]{Training}  &  18 & 12 & 14 & 11 & 11 & 16 & 12 & 10 & 11 & 8 & 11 & 7 & 11\\
    \makecell[l]{Testing}   & 42 & 118 & 11 & 103 & 44 & 40 & 83 & 20 & 15 & 7 & 20 & 8 & 9\\ 
    \makecell[l]{Labeled}  & 22& 22 & 11 & 22 & 22 & 22 & 22 & 20 & 15 & 7 & 20 & 8 & 9\\
      
    \bottomrule
    \end{tabular}

\end{table}

Our MRC algorithm was compared with two other state-of-the-art algorithms, WSSD~\cite{zhu2010image} and CRD~\cite{wagner2021real}. To show that linear regression is suitable for robot application, we compare it with a typical nonlinear regression method named Gaussian Process Regression (GPR)\footnote{Details can be found in the supplementary material.}\cite{williams2006gaussian}. All the hyperparameters were pre-tuned to guarantee the best performance. All experiments were conducted on a commercial desktop with Intel Core i5-10500 CPU at 3.10GHz with 16Gb memory. The hyperparameter of our proposed MRC $\rho_{th}$ was set as $0.9$ in our experiments.

\begin{figure*}[ht]
    \centering
    \includegraphics[width=17cm]{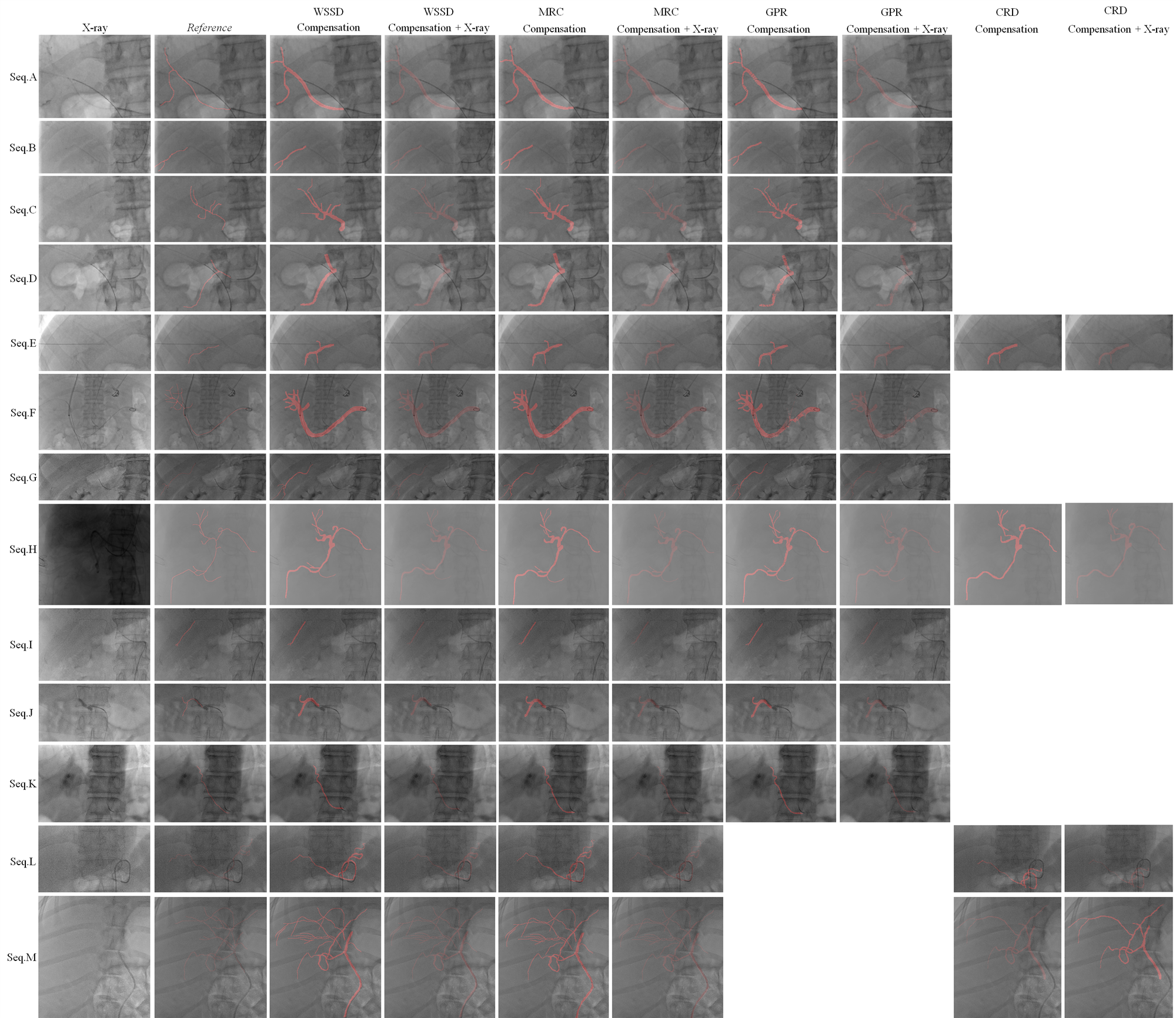}
    \caption{Visualization of vascular respiratory motion compensation. Each row represents one sample frame. The first column is the original image, the second column is the \textit{reference} labeled by experts, and the remaining eight columns are the results of WSSD, MRC, GPR, and CRD algorithms. For every algorithm, the compensation column uses bright red to emphasize the vascular mask, and the compensation + X-ray column uses dark red to observe accuracy.}
    \label{fig:visualization}
\end{figure*}

\subsection{Evaluation criteria}

Vascular MRC algorithms can be evaluated in terms of time and accuracy. The accuracy can be evaluated qualitatively by experienced physicians and quantitatively with the average Euclidean distance between predicted and \textit{reference} frames. In this paper, we adopt ratio $\mathrm{R}$ and mean Euclidean distance $\mathrm{MD}$ to quantitatively evaluate algorithms accuracy, which are calculated by 

\begin{equation}
    \mathrm{R} = \frac{\vert{\mathbf{M}_{gt}\cap{\mathcal{M}(\mathbf{M}_{r}, \Tilde{\mathbf{F}}_{q}^{v})}}\vert}{\vert{\mathbf{M}_{gt}}\vert},
    \label{ratio}
\end{equation}

\begin{equation}
    \mathrm{MD} = \frac{1}{\mathrm{N}_{p}} \sum_{m=1}^{\mathrm{N}_{p}}\Vert g(\mathbf{M}_{gt})(m) - g(\mathcal{M}(\mathbf{M}_{r}, \Tilde{\mathbf{F}}_{q}^{v}))
   (m)\Vert_{2},
    \label{distance}
\end{equation}

\noindent where $\mathbf{M}_{gt}$ is the labeled \textit{reference} centerline, function $\mathcal{M}$ maps $\mathbf{M}_{r}$ onto live image $\mathbf{R}_{q}$ using motion $\Tilde{\mathbf{F}}_{q}^{v}$, $g$ extracts centerline point coordinates of mask, $\mathrm{N}_{p}$ is the number of centerline points in $\mathbf{M}_{gt}$.

\subsection{Experiment Results}

We conducted vascular compensation experiments on 13 sequences. It should be noted that CRD~\cite{wagner2021real} was tested on 4 sequences because data input required the image to contain the liver's top. For Seq.L and Seq.M, GPR is not conducted because of the very long training time caused by large data. What's more, our MRC's error is low enough (mean $MD=0.652 mm$ and mean $MD=0.523 mm$). Even if GPR's error on Seq.L and Seq.M is zero, the average accuracy of GPR is lower than our MRC.

\subsubsection{Accuracy}

\begin{table*}[]
    \renewcommand{\arraystretch}{1.} 
    \setlength{\tabcolsep}{7.pt} 
    \caption{The mean score of two clinicians grading according to the visualization roadmap. 5 is a perfect score. Seq.A-Seq.M represents different sequences.}
    \label{tab:score}
    \centering
   
    \begin{tabular}{l c c c c c c c c c c c c c c}
   \toprule
   
     & Seq.A & Seq.B & Seq.C & Seq.D & Seq.E & Seq.F & Seq.G & Seq.H & Seq.I & Seq.J & Seq.K & Seq.L & Seq.M & mean\\

    \midrule
   WSSD  &  3 & 2 & 3.5 & 3.5 & 4.5 & 2 & 2 & 3.5 & 3 & 2.5 & 3 & 4 & 3.5 & 3.077\\
    MRC   & 4 & 3 & 4 & 3.5 & 4.5 & 4 & 3 & 4.5 & 4 & 4.5 & 4 & 4.5 & 4 & 3.962\\ 
    GPR & 2.5 & 2 & 3 & 2 & 4 & 1.5 & 1.5 & 4 & 3.5 & 3 & 3.5 & - & - & 2.773\\
    CRD  & - & - & - & - & 2.5 & - & - & 0.1 & - & - & - & 0.1 & 0.1 & 0.7 \\
      
    \bottomrule
    \end{tabular}

\end{table*}

\begin{table*}[]
    \renewcommand{\arraystretch}{1.} 
    \setlength{\tabcolsep}{2.5pt} 
    \caption{The model learning time[s] for different sequences and different methods. Seq.A-Seq.M represents different sequences.}
    \label{tab:train time}
    \centering
   
    \begin{tabular}{l c c c c c c c c c c c c c}
    \toprule
   
     & Seq.A & Seq.B & Seq.C & Seq.D & Seq.E & Seq.F & Seq.G & Seq.H & Seq.I & Seq.J & Seq.K & Seq.L & Seq.M\\

    \midrule
    WSSD  &  110.127 & 82.646 & 121.894 & 68.250 & 272.435 & 376.654 & 593.059 & 232.944 & 262.118 & 117.140 & 150.492 & 152.328 & 235.327\\
    MRC   & \bf{0.077} & \bf{0.029} & \bf{0.041} & \bf{0.029} & \bf{0.060} & \bf{0.404} & \bf{0.221} & \bf{0.071} & \bf{0.076} & \bf{0.044} & \bf{0.039} & \bf{0.044} & \bf{0.089}\\ 
    
    GPR & 10935.236 & 4978.496 & 13922.305 & 7642.906 & 28351.750 & 268172.254 & 194248.282 & 29759.035 & 5347.611 & 15609.621 & 10876.458 & - & - \\
    CRD  & - & - & - & - & 12.641 & - & - & 10.755 & - & - & - & 5.105 & 18.140 \\
      
    \bottomrule
    \end{tabular}

\end{table*}

\begin{table*}[]
    \renewcommand{\arraystretch}{1.} 
    \setlength{\tabcolsep}{6pt} 
    \caption{The mean of time[s] of each frame to predict vascular motion on 13 sequences. Seq.A-Seq.M represents different sequences.}
    \label{tab:test time}
    \centering
   
    \begin{tabular}{l c c c c c c c c c c c c c c}
    \toprule
         
     & Seq.A & Seq.B & Seq.C & Seq.D & Seq.E & Seq.F & Seq.G & Seq.H & Seq.I & Seq.J & Seq.K & Seq.L & Seq.M & Mean\\

    \midrule
    WSSD  &  0.028 & 0.050 & 0.041 & 0.033 & 0.134 & 0.130 & 0.355 & 0.091 & 0.124 & 0.090 & 0.093 & 0.156 & 0.143 & 0.116 \\
    MRC   & \bf{0.012} & \bf{0.019} & \bf{0.008} & \bf{0.005} & 0.030 & \bf{0.124} & \bf{0.046} & 0.016 & \bf{0.011} & \bf{0.035} & \bf{0.029} &  \bf{0.040} & 0.155 & 0.032\\ 
    GPR & 17.265 & 7.785 & 22.405 & 12.852 & 47.526 & 520.735 & 328.753 & 50.823 & 8.835 & 27.107 & 17.027 & - & - & 109.543 \\
    CRD  & - & - & - & - & \bf{0.002} & - & - & \bf{0.001} & - & - & - & 0.002 & 0.002 & 0.002 \\
      
    \bottomrule
    \end{tabular}

\end{table*}

\begin{figure*}[htbp]
    \centering
    \includegraphics[width=17cm]{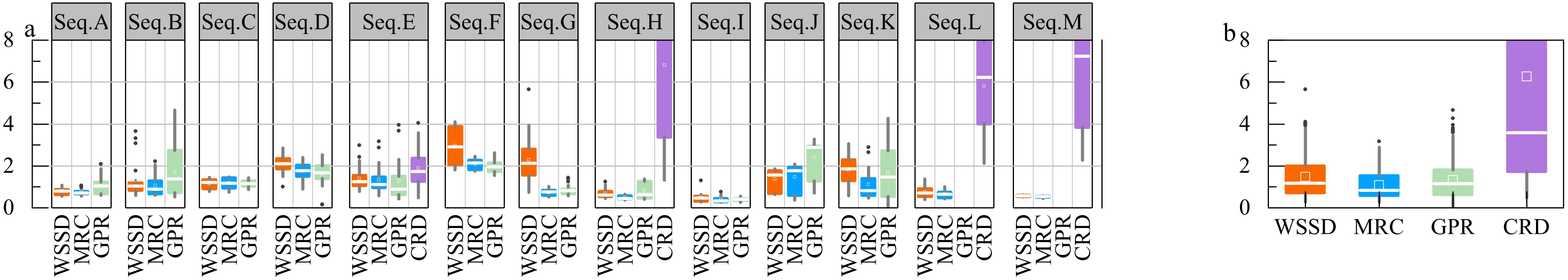}
    \caption{The box chart of mean Euclidean distance[mm] on 13 sequences. (a) shows $\mathrm{MD}$ on 13 sequences respectively, Seq.A-Seq.M represents different sequences. (b) shows the overall results on all 13 sequences. It should be noted that the overall results of CRD are on 4 sequences.}
    \label{fig:distance}
\end{figure*}

\begin{figure*}
    \centering
    \includegraphics[width=17cm]{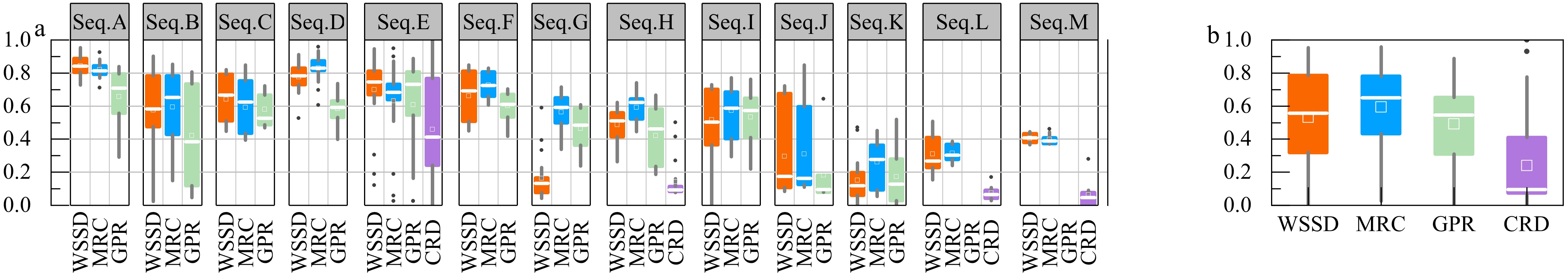}
    \caption{The box chart of value of ratio $\mathrm{R}$ on 13 sequences. (a) shows $\mathrm{R}$ on 13 sequences respectively, Seq.A-Seq.M represents different sequences. (b) shows the overall results on all 13 sequences. It should be noted that the overall results of CRD are on 4 sequences.}
    \label{fig:ratio}
\end{figure*}

To assess the compensation accuracy, two clinicians scored them using the compensated roadmaps. Table \ref{tab:score} shows the mean score, which indicates that MRC possesses the best visualization for clinicians. Fig.~\ref{fig:visualization} shows some sample results.\footnote{We strongly recommend readers watch the video to appreciate the results.} It should be noted that in order to provide better visualization, only the image of the vascular region is shown. As Fig.~\ref{fig:visualization} indicates, there are cases where the results of WSSD, GPR, and MRC look the same, such as for sequences A, C, D, E, H, I, and M. In some cases, MRC works better than WSSD and GPR, such as in sequences B, F, and J. For sequences G, K, L, and M, MRC and GPR look the same and better than WSSD. CRD is worse than WSSD and MRC in sequences H, L, and M. According to the frame image shown in the figure, the result of CRD is the same as WSSD and MRC for sequences E. However, in some frames of sequence E, the visualization results are worse than that of WSSD and MRC. And its overall results vary considerably, which can be concluded from Fig.~\ref{fig:distance} and Fig.~\ref{fig:ratio}. Although the error variance of CRD can be decreased by increasing the number of training X-ray images, patients suffer from higher doses of X-ray.

Results of the mean Euclidean distance on 13 sequences are shown in Fig.~\ref{fig:distance}. It can be seen that $\mathrm{MD}$ of CRD is dispersed, and its mean and median are higher. For sequences A, B, F, G, H, I, K, L, and M MRC performs best on $\mathrm{MD}$. For sequences C, D, and E, GPR possesses minimum error on $\mathrm{MD}$. For sequences J, WSSD performs best on $\mathrm{MD}$. Fig.~\ref{fig:distance} (b) shows that MRC achieves the best accuracy.

Fig.~\ref{fig:ratio} shows the results of ratio $\mathrm{R}$. 1 means the best prediction, and 0 indicates failure. It can be seen that $\mathrm{R}$ of CRD is lower than the other three algorithms. For sequences B, D, F, G, H, I, K, and L, MRC possesses the highest $\mathrm{R}$. For sequences A, C, E, J, and M, WSSD performs best on $\mathrm{R}$ than the other three algorithms. In general, our proposed MRC performs best on $\mathrm{R}$ as shown in Fig.~\ref{fig:ratio} (b). The error of the sparse corners alignment algorithm varies with the image and can not be estimated, which makes GPR's accuracy vary with the accuracy of the additive Gaussian noise prior. To sum up, our MRC performs best on accuracy, which benefits from a reasonable motion model in that we consider the difference of motion in different positions incurred by respiration.

\subsubsection{Time Consumption}

Real-time performance is critical for clinical usage. Both learning time consumption and motion-prediction time consumption are tested. The model learning time is shown in Table \ref{tab:train time}. The proposed MRC performs well on model learning time. The mean of each frame predicting vascular motion time is shown in Table \ref{tab:test time}. It demonstrates that the prediction time of MRC is less than WSSD and GPR on all 13 sequences. Although GPR can be accelerated~\cite{ranganathan2010online, kabzan2019learning}, its time cannot exceed linear regression. For sequences E, H, L, and M, the prediction time of CRD is less than MRC, but its robustness and accuracy are limited. In the meantime, application scenarios of CRD need the FOV of the image including the top of the liver. In summary, our MRC performs best on learning time and second best on prediction time because of relatively simple modeling and sparse point tracker. CRD is the fastest approach since it estimates the respiratory state from the image to predict motion. However, it is not robust if the respiration pattern changes.

\subsubsection{Ablation Study}
Ablation experiments were conducted to verify the efficiency of our GOF and sparse optical flow. The results are shown in Table \ref{tab:ablation}. Our MRC is tested with and without GOF. We also test the choice of dense and sparse optical flow. Results indicate that sparse optical flow can significantly reduce time consumption. GOF can improve the accuracy of sparse optical flow prediction with some time consumption. Still, GOF is basically ineffective for dense optical flow prediction due to the more accurate dense optical flow algorithm. Thus, sparse motion flow with GOF possesses compromised accuracy and time.

\begin{table}
    \renewcommand{\arraystretch}{1.} 
    \setlength{\tabcolsep}{2.5pt} 
    \caption{Results of ablation experiments in accuracy and time. Sparse and dense represent sparse motion flow and dense motion flow.}
    \label{tab:ablation}
    \begin{tabular}{lcccc}
    \toprule
                   & MD{[}mm{]} & R     & \multicolumn{1}{c}{prediction time{[}s{]}} & Learning time{[}s{]} \\ 
    \midrule
sparse with GOF    & \bf{1.086}     & \bf{0.595} & 0.032                                    & 0.092              \\
sparse without GOF & 1.565      & 0.432 & \bf{0.012}                                      & \bf{0.091}                \\
dense with GOF     & 1.192     & 0.585 & 10.587                                        & 26.691              \\
dense without GOF  & 1.192      & 0.587 & 4.842                                       & 21.524               \\ \bottomrule
    \end{tabular}

\end{table}

\subsection{Limitations \& Future Works}

Our proposed MRC comes with three drawbacks.
First, only one frame contrasted image vascular mask is used to map onto the live image, the mapped vascular mask size may be small. Future works will conduct the fusion multi-frames contrasted images of vascular information to enrich the vascular mask. 
Secondly, the predicted vascular motion flow is not smooth enough due to the sparse flow of vascular. Although the local region is not smooth, it does not affect physicians' overall judgment.
Lastly, it cannot handle the scenario with heart-beat because of poor image quality, leading to the error in optical flow\footnote{The last video clip shows the failure case.}. We plan to utilize optical flow based on deep learning to compute flow motion.

\section{Conclusion}

We propose MRC to conduct vascular respiratory motion compensation in real-time to predict vascular on the live fluoroscopic image with invisible vascular. Based on the linear correlation between vascular motion flow and non-vascular motion flow, multi-frame contrasted images are used to train a motion-related model. In the prediction stage, predictions from non-vascular points are refined with GOF. The proposed method was tested on 13 in-vivo vascular intervention fluoroscopic sequences. Results show that the proposed method achieves a compensation accuracy of $1.086~mm$ in $0.032~s$. Our approach provides a practical real-time solution to assist physicians/robots in vascular interventions. This work is in the process of commercialization by the company United Imaging of Health Co., Ltd.

\balance
{\small
		\bibliographystyle{IEEEtranN}
		\bibliography{bib/strings-abrv,bib/ieee-abrv,main}
	}

\end{document}


\title{Respiratory motion compensation based on Gaussian process regression}
\maketitle

The aim of this supplementary material is to validate that our linear regression algorithm is enough to handle relationship modelling. Although machine learning methods may substitute linear regression, they are slow and sensitive to parameter tuning as shown in our research. This supplementary material cites GPR as an example to learn the nonlinear relationship between vascular and non-vascular motion. Since no research paper has been published (the idea is straightforward), technical details of our implementation are presented in this note for further re-implementations.\par

We would like to emphasize again that linear regression has similar accuracy ($1.086mm$ compared with $1.330mm$), requires almost no parameter tuning (only Pearson threshold and not sensitive), and is much faster than machine learning methods. It is the most suitable algorithm for real-world robot-assisted vascular interventions among all candidate approaches. \par


\section{Gaussian process regression}

In the following, we briefly introduce Gaussian process regression(GPR). In regression, there is a function $\mathbf{f}$ we are trying to model given observed data points $(\mathbf{x_{k}},\mathbf{y_{k}}),k=1,...,\mathrm{N}$ (training dataset) from the unknown function $\mathbf{f}$.

\begin{equation}
    \mathbf{y_{k}} = \mathbf{f}(\mathbf{x_{k}}) + \mathbf{\sigma_{n}},
\end{equation}
where $\mathbf{\sigma_{n}}$ is the measurement noise. In Gaussian process regression, the regression function is modeled by a multivariate Gaussian as
\begin{equation}
    P(\mathbf{f}|\mathbf{X}) \sim \mathcal{N}(\mathbf{f}|\boldsymbol{\mu},\mathbf{K}),
    \label{GPR function}
\end{equation}
\noindent where $\mathbf{X} = [\mathbf{x}_{1},...,\mathbf{x}_{N}]$, $\mathbf{f} = [\mathbf{f}(\mathbf{x}_{1}), ..., \mathbf{f}(\mathbf{x}_{N})]$, $\boldsymbol{\mu}=[m(\mathbf{x}_{1}), ..., m(\mathbf{x}_{N})]$ and $K_{i,j} = k(\mathbf{x_{i}},\mathbf{x_{j}})$. $m$ represents the mean function, and $k$ represents a positive definite kernel function. For the Radial Basis Function (RBF) kernel function, $k(\mathbf{x}_{i}, \mathbf{x}_{j})=cexp\{-(\Vert \mathbf{x}_{i}-\mathbf{x}_{j} \Vert_{2}) / (2\eta^{2}) \}$. There are hyperparameters $\boldsymbol{\theta}=(c, \eta)$ in RBF, which can be learned by maximizing the marginal log-likelihood

\begin{equation}
    p(\mathbf{Y}|\mathbf{X},\boldsymbol{\theta}) = - \frac{1}{2}\log|\mathbf{K}+\boldsymbol{\sigma}_{n}^{2}I|-\frac{1}{2}\mathbf{Y}^\top(\mathbf{K}+\boldsymbol{\sigma}_{n}^{2}I)^{-1}\mathbf{Y}-\frac{N}{2}\log2\pi,
    \label{hyperpara optimize}
\end{equation}

\noindent where $\mathbf{Y}=[\mathbf{y}_{1},...,\mathbf{y}_{N}]$, $I$ is the identity matrix of same dimensions as $\mathbf{K}$.

We make predictions at new points $\mathbf{X}_{*}$ as $\mathbf{f}({\mathbf{X}_{*}})$. The joint
distribution of the observed values and the function values at new testing
points are

\begin{equation}
    \left(
    \begin{array}{c}
        \mathbf{Y}  \\
         \mathbf{f}_{*}
    \end{array} 
    \right)
    \sim \mathcal{N}
    \left(
    \left[
    \begin{array}{c}
    m(\mathbf{X}) \\
    m(\mathbf{X}_{*})
    \end{array}
    \right],
    \left[
    \begin{array}{cc}
      \mathbf{K}+ \sigma_{n}^{2}I   & \mathbf{K}_{*} \\
       \mathbf{K}_{*}^\top  & \mathbf{K}_{**}
    \end{array}
    \right]
    \right),   
\end{equation}

\noindent where $\mathbf{K}=K(\mathbf{X}, \mathbf{X})$, $\mathbf{K}_{*}=K(\mathbf{X}, \mathbf{X}_{*})$, $\mathbf{K}_{**}=K(\mathbf{X}_{*}, \mathbf{X}_{*})$.

By deriving the conditional distribution, we get the predictive equations for GPR as
\begin{equation}
    \mathbf{f}_{*}|\mathbf{X},\mathbf{Y},\mathbf{X}_{*} \sim \mathcal{N}(\mathbf{\overline{f}}_{*}, \mathrm{cov}(\mathbf{f}_{*})),
    \label{GPR predicting}
\end{equation}

\noindent where $\mathbf{\overline{f}}_{*}=\mathbf{K}_{*}^\top[\mathbf{K}+\boldsymbol{\sigma}_{n}^{2}I]^{-1}\mathbf{Y}$, $\mathrm{cov}(\mathbf{f}_{*})=\mathbf{K}_{**}-\mathbf{K}_{*}^\top[\mathbf{K}+\boldsymbol{\sigma}_{n}^{2}I]^{-1}\mathbf{K}_{*}$. 

\section{model establishing based on GPR}

Like MRC, respiratory motion compensation based on GPR considers that there is a correlation between vascular motion and non-vascular motion, and GPR is used to find the relationship between vascular motion and non-vascular motion. The process of establishing the motion-related model based on GPR is shown in Algorithm \ref{gpr training}(Variable definitions are the same as in the main text). The additive Gaussian noise prior is set as $\boldsymbol{\sigma}_{n}=0.01$ according to the result of experiments.

\begin{algorithm}[]
    \small 
    \caption{Motion-related model establishing based on GPR process}
    \label{gpr training}
    \begin{algorithmic}[1]
    \Require  reference frame $\mathbf{I}_{r}$, contrasted sequences $\mathbf{I}$

    \Ensure $\mathrm{N}_{v} \times \mathrm{N}_{n} \times 2$ GPR models $\mathbf{G}$
    
    \State Calculate training vascular motion flows $ \{ \mathbf{D}_{1}^{v}, \mathbf{D}_{2}^{v}, ..., \mathbf{D}_{k}^{v} \}$ and training non-vascular motion flows $ \{ \mathbf{D}_{1}^{n}, \mathbf{D}_{2}^{n}, ..., \mathbf{D}_{k}^{n} \}$
    
    \For{each vascular corner $i \in [1, \mathrm{N}_{v}]$}
    \For{each non-vascular corner $j \in [1, \mathrm{N}_{n}]$}
    
    \State modeling the regression function between $\mathbf{Y}_{i}|_{x}\in \mathbb{R}^{1 \times \mathrm{k}}$ and $\mathbf{X}_{j}|_{x} \in \mathbb{R}^{1 \times k}$ according \eqref{GPR function} and \eqref{hyperpara optimize} as $\mathbf{G}_{i,j,1}$\footnotemark[1]

    \State modeling the regression function between $\mathbf{Y}_{i}|_{y}\in \mathbb{R}^{1 \times \mathrm{k}}$ and $\mathbf{X}_{j}|_{y} \in \mathbb{R}^{1 \times k}$ according \eqref{GPR function} and \eqref{hyperpara optimize} as $\mathbf{G}_{i,j,2}$
       
    \EndFor
    \EndFor
        
    \end{algorithmic}
\end{algorithm}

\footnotetext[1]{The toolkit scikit-learn(https://scikit-learn.org/stable/index.html) is used to implement GPR. }

\section{motion predicting based on GPR}

For a live image without the contrast agent $\mathbf{R}_{q}$, vascular motion between $\mathbf{R}_{q}$ and $\mathbf{I}_{r}$ can be predicted utilizing GPR models $\mathbf{G}$ and non-vascular corners motion $\mathbf{F}_{q}^{n}$. The variance of $\mathbf{G}_{i,j}$ on the test data is denoted as $\mathbf{v}_{i}^{(j)} \in \mathbb{R}^{1 \times 2}$. Define $\mathbf{V}_{i}=[\mathbf{v}_{i}^{(1)}, ..., \mathbf{v}_{i}^{(\mathrm{N}_{n})}]^\top \in \mathbb{R}^{\mathrm{{N}_{n}} \times 2}$. The $i$th vascular corner weight matrix of each non-vascular is denoted as $\mathbf{W}_{i}=[\mathbf{w}_{i}^{(1)}, ..., \mathbf{w}_{i}^{(\mathrm{N}_{n})}]^\top \in \mathbb{R}^{\mathrm{{N}_{n}} \times 2}$. In motion compensation based on the motion-related model, each vascular motion prediction $\hat{\mathbf{f}}_{q}^{v}(i)$ is decided by the linear combination of $\hat{\mathbf{p}}_{i}^{(j)}$. Therefore, the variance of $\hat{\mathbf{f}}_{q}^{v}(i)$ can be estimated by 

\begin{equation}
    \overline{\mathbf{v}}|_{x}=\mathbf{W}_{i}|_{x}^\top \cdot \mathbf{V}_{i}|_{x},
    \overline{\mathbf{v}}|_{y}=\mathbf{W}_{i}|_{y}^\top \cdot \hat{\mathbf{V}}_{i}|_{y}.
    \label{combination variance}
\end{equation}

\noindent $\overline{\mathbf{v}}$ represents the uncertainty about the predicting $\hat{\mathbf{f}}_{q}^{v}(i)$, which can be used to delete some unreliable predicting to improve the accuracy of motion compensation. The process of vascular motion predicting is shown in Algorithm \ref{gpr test}.

\begin{algorithm}
    \small
    \caption{Predicting vascular motion flows based on GPR.}
    \label{gpr test}
    \begin{algorithmic}[1]
        \Require
        GPR models $\mathbf{G}$, live image $\mathbf{R}_{q}$, reference frame $\mathbf{I}_{r}$
        
        \Ensure
        predicted vascular motion flows $\hat{\mathbf{F}}_{q}^{v} \in \mathbb{R}^{\mathrm{N}_{v} \times 2}$ between $\mathbf{R}_{q}$ and $\mathbf{I}_{r}$

        \State Calculate non-vascular motion flows $\mathbf{F}_{q}^{n} \in \mathbb{R}^{\mathrm{N}_{n} \times 2}$ between $\mathbf{R}_{q}$ and $\mathbf{I}_{r}$

        \For{each vascular corner $i \in [1, \mathrm{N}_{v}]$}

        \For{each non-vascular corner $j \in [1, \mathrm{N}_{n}]$}

        \State for the new data $\mathbf{f}_{q}^{n(j)}|_{x}$, its vascular motion predicting $\hat{\mathbf{p}}_{i}^{(j)}|_{x}$ and variance $\mathbf{v}_{i}^{(j)}|_{x}$ can be calculated according to \eqref{GPR predicting}
        
        \State for the new data $\mathbf{f}_{q}^{n(j)}|_{y}$, its vascular motion predicting $\hat{\mathbf{p}}_{i}^{(j)}|_{y}$ and variance $\mathbf{v}_{i}^{(j)}|_{y}$ can be calculated according to \eqref{GPR predicting}

        \State $\mathbf{w}_{i}^{(j)}|_{x} = 1 / (\mathbf{v}_{i}^{(j)}|_{x}$), $\mathbf{w}_{i}^{(j)}|_{y} = 1 / (\mathbf{v}_{i}^{(j)}|_{y})$
        
        \EndFor
        
        \State Normalize weight $\mathbf{W}_{i}|_{x} = \frac{\mathbf{W}_{i}|_{x}}{\sum_{j=1}^{\mathrm{N}_{n}}\mathbf{w}_{i}^{(j)}|_{x}}$, $\mathbf{W}_{i}|_{y} = \frac{\mathbf{W}_{i}|_{y}}{\sum_{j=1}^{\mathrm{N}_{n}}\mathbf{w}_{i}^{(j)}|_{y}}$

        \State Predict vascular motion flow $\hat{\mathbf{f}}_{q}^{v}(i)|_{x}=\mathbf{W}_{i}|_{x} \cdot \hat{\mathbf{P}}_{i}|_{x}$, $\hat{\mathbf{f}}_{q}^{v}(i)|_{y}=\mathbf{W}_{i}|_{y} \cdot \hat{\mathbf{P}}_{i}|_{y}$

        \State /*   Delete   predicting with large uncertainty   */ 
        \State Estimate variance $\overline{\mathbf{v}}$ of $\hat{\mathbf{f}}_{q}^{v}(i)$ according to \eqref{combination variance}  
        \If{$\overline{\mathbf{v}} >  \overline{\mathbf{v}}_{th}$}
        \State Delete the predicting $\hat{\mathbf{f}}_{q}^{v}(i)$ 

        \EndIf
      
        \EndFor
  
    \end{algorithmic}
    
\end{algorithm}